\documentclass[runningheads]{llncs}
\usepackage{graphicx}
\usepackage{float}
\usepackage{xcolor}
\usepackage{cite}
\usepackage{array}
\usepackage{multirow}
\usepackage{enumitem}
\usepackage{hyperref}

\begin{document}
\title{Forecasting Health and Wellbeing for Shift Workers Using Job-role Based Deep Neural Network \thanks{Supported by National Science Foundation \# 1840167 and Japan Agency for Medical Research and Development.}}
\titlerunning{Forecasting Health and Wellbeing for Shift Workers}
\author{Han Yu\inst{1} \and
Asami Itoh\inst{2} \and
Ryota Sakamoto\inst{2} \and
Motomu Shimaoka\inst{2} \and
Akane Sano\inst{1} }

\authorrunning{H. Yu et al.}
\institute{Rice University, Houston TX 77005, USA  \\
\email{Han.Yu and akane.sano@rice.edu}\\
\and Mie University, Mie 514-8507, Japan \\
\email{amasui and sakamoto@clin.medic.mie-u.ac.jp \\ shimaoka@doc.medic.mie-u.ac.jp}\\
}

\maketitle              % typeset the header of the contribution
\begin{abstract}
Shift workers who are essential contributors to our society, face high risks of poor health and wellbeing. To help with their problems, we collected and analyzed physiological and behavioral wearable sensor data from shift working nurses and doctors, as well as their behavioral questionnaire data and their self-reported daily health and wellbeing labels, including alertness, happiness, energy, health, and stress. We found the similarities and differences between the responses of nurses and doctors. According to the differences in self-reported health and wellbeing labels between nurses and doctors, and the correlations among their labels, we proposed a job-role based multitask and multilabel deep learning model, where we modeled physiological and behavioral data for nurses and doctors simultaneously to predict participants' next day's multidimensional self-reported health and wellbeing status. Our model showed significantly better performances than baseline models and previous state-of-the-art models in the evaluations of binary/3-class classification and regression prediction tasks. We also found features related to heart rate, sleep, and work shift contributed to shift workers' health and wellbeing.
\end{abstract}
\section{Introduction}
\label{intro}
Around 20\% of the workforce in the world involves in shift work \cite{wright2013shift}. Their irregular shift work brings a high risk of poor health and wellbeing. For example, shift work disrupts workers' circadian rhythms and causes problems such as sleep disorder and insomnia \cite{garbarino2002sleepiness}. In addition to the sleep issues, decreased alertness levels were found in healthy shift workers \cite{ganesan2019impact}, which could lead to occupational errors and accidents. Previous studies also showed the potential associations between shift work and pathological disorders such as fatigue, gastrointestinal malfunction \cite{knutsson2003health}, and an increased risk of colorectal cancer in night shift nurses \cite{schernhammer2003night}. 
Moreover, more adverse mental health outcomes, emotional exhaustion, and burnout were observed in shift workers compared to daytime workers \cite{srivastava2010shift, courtney2010caring, vogel2012effects, wisetborisut2014shift, jamal2004burnout}. In health care domain, physician burnout is estimated to cost 4.6 billion USD per year \cite{han2019estimating}.

To support shift worker's health and wellbeing, monitoring and predicting their day-to-day health and wellbeing trajectories and providing aids to help them prepare for challenging situations might be useful. Besides, mobile devices, such as smartphones and wearable sensors, have become parts of people's daily life, and have been used to detect and predict self-reported health and wellbeing with the help of machine learning models\cite{Previous-cla, Previous-reg, Previous-reg2, stress-cla1, likamwa2013moodscope, umematsu2019improving}. These previous works targeted health and wellbeing detection or prediction as binary classification \cite{stress-cla1, Previous-cla}, 3-class classification \cite{Previous-reg2, Muaremi-stress}, and regression tasks \cite{Previous-reg, Previous-reg2,asselbergs2016mobile}. Some of these works developed personalized models by taking participants' demographic information into account \cite{Previous-cla} or fine-tuning general models to specific users \cite{yu2020passive}. Correlations among self-reported multi-dimensional labels - including subjective mood, health, and stress- were also used in building multilabel neural network models \cite{Previous-cla}. In addition, there are some prior works in monitoring shift workers using wearable sensors. Feng \textit{et al.} extracted a behavioral consistency feature from shift worker wearable data and estimated anxiety levels with an accuracy of 57.8\% in binary classification. Mulhall \textit{et al.} used sensors integrated in the vehicles to monitor shift workers' eye blinking as a marker of alertness \cite{mulhall2020pre} while driving. Actigraphy has been also used widely for studying sleep for shift work nurses  \cite{geiger2012sleep, kato2012sleepiness}.

Although these previous works have achieved promising results, there is no work to thoroughly monitor and analyze different job types of shift workers' multidimensional wellbeing and forecast them using machine learning. Furthermore, the models developed previously considered the heterogeneity among participants and correlation among wellbeing labels separately; however, since these two characteristics ubiquitously co-exist, modeling them simultaneously for different job types of shift workers might improve prediction model performance.

In this work, we collected physiological and behavioral data from hospital shift workers, then we developed machine learning models to predict their next day's wellbeing in binary/3-class classifications and regression tasks. 
We also verified the rationale of leveraging job role information and multi wellbeing labels simultaneously in the models by analyzing the data. Then, we proposed a multitask multilabel deep learning model that leveraged job role information and correlations among self-reported health and wellbeing labels.

Our contributions can be summarized as:
(i) we collected physiological and behavioral data from hospital shift workers, including nurses and doctors,
(ii) we analyzed their physiological and behavioral patterns and found similarities and differences,
(iii) we developed a multitask multilabel deep learning model to predict participants' near future wellbeing using wearable sensor, surveys, their job role information, and correlations among wellbeing labels. The details of our proposed model structure, implementation and hyper-parameter information are shared on:  \hyperlink{https://github.com/comp-well-org/multitask-multilabel-wellbeing-prediction}{\textit{https://github.com/comp-well-org/multitask-multilabel-wellbeing-prediction}}.

\section{Related Work}
There are numerous studies on shift workers' health and wellbeing. Heath \textit{et al.} collected survey data from shift work nurses, and applied statistical analysis in exploring the association among their work shift types, sleep, mood, and diet \cite{heath2019associations}. They showed that shift work was significantly negatively related to shit workers' diet, sleep efficiency, and stress levels. Similarly, Books \textit{et al.} analyzed questionnaire data from shift-working nurses and showed an increased risk of sleep deprivation, family stressors, and mood changes due to the night work shift \cite{books2017night}.

In addition, with the rapid development of mobile devices and mobile applications, objective data from wearables and smartphones have been used for studying shift workers. 
For example, Pereira \textit{et al.} collected wearable accelerometer data from hospital shift workers and detected 4 levels of their physical activity intensity with an 83\% accuracy score \cite{pereira2018physical}. Feng \textit{et al.} used wearable devices to collect physiological data from shift work nurses for ten weeks and applied a clustering method for extracting behavioral consistency, which intuitively captures unique behavioral patterns between different groups of nurses \cite{feng2020modeling}. They further found that behavioral consistency can help predict self-reported work behaviors and anxiety levels.
In another work, Feng \textit{et al.} analyzed physiological and indoor location data from nurses with Fitbit wrist-wearable devices and Bluetooth hubs \cite{feng2020modeling1}. They extracted mutual information features and demonstrated the dependency between an individual’s movement patterns and physiological responses.
% \textcolor{red}{need to introduce https://ieeexplore.ieee.org/document/9054493
% https://ieeexplore.ieee.org/document/9054307
% }

Machine learning models have been designed for detecting or predicting health and wellbeing using mobile and sensor data. 
For example, Bogomolov \textit{et al.} developed daily stress detection algorithms based on five-month-long weather, mobile phone data (e.g., calls, SMS, and screen usage), and personality survey data from 117 participants \cite{stress-cla1}. They obtained stress detection accuracy up to 72\% in binary classification tasks. 
%Besides the classification models, regression models have also been developed \cite{likamwa2013moodscope,asselbergs2016mobile}. 
In Moodscope paper, mood (1: negative to 5: positive) was detected with the best mean squared error of 0.229 using the data from the mobile phone and a personalized linear regression model \cite{likamwa2013moodscope}. Similarly, Asselbergs \textit{et al.} detected the current mood using mobile phone data with a mean squared error of 0.15 out of -2 to 2 mood scale \cite{asselbergs2016mobile}. For further improving the model performance, Taylor \textit{et al.} developed a multitask machine learning model to predict high/low self-reported stress, mood, and health and separately used (i) the demographic information such as gender and personalities of participants and (ii) correlations among labels \cite{Previous-cla}. This work also inspired us to use the combination of job role information and label correlations. In this work, we study the differences in daily self-reported health and wellbeing, physiology, and behavior between nurses and doctors, and focus on estimating shift workers' health and wellbeing using the data from mobile sensors and surveys and job-role based deep learning models.

\section{Methods}
\subsection{Data Collection}
\label{fitbit_data}

Two hundred and forty-one days of multi-modal data were collected from 14 shift workers, including 10 nurses (one male) and 4 doctors (all males) in a hospital in Japan. The average age of all participants was 31.4 years old, with a standard deviation (SD) of 4.2. For each study day, participants wore a Fitbit wristwatch (Fitbit Charge 3) for monitoring their physiological and behavioral activities such as heart rate, sleep, and step counts. The data sampled every 1 minute was downloaded from the Fitbit server for data analysis and modeling. In addition, participants filled out daily morning and evening questionnaires to record their behavioral activities, including sleep, work schedule, and caffeinated drinks, alcohol \& drug intake.

Self-reported health and wellbeing labels - including alertness, happiness, energy, health, and stress - were also collected in the morning questionnaire using 0 to 100 scales, with 0 to the most negative and 100 being the most positive (sleepy-alert, sad-happy, sluggish-energetic, sick-healthy, stressed-calm). 

\subsection{Features}
We calculated the following features from the Fitbit data and daily questionnaires:
\subsubsection{Heart Rate} Heart rate and heart rate variability are related to work stress \cite{vrijkotte2000effects} and mood \cite{shapiro2001striking}. Based on heart rate collected from Fitbit sensor every 1 minute, we computed features including daily mean, standard deviation (SD), and entropy of heart rates. We computed sample entropy of heart rate, which represented the self-similarity of a sequence and has been used in physiological time-series data analysis \cite{richman2000physiological}. To calculate the sample entropy, we first need to set an embedding dimension $m$. Using the given $m$, our sequence $X$ with length $N$ can be divided into $N-m+1$ sliding windows \{$X_1$,...,$X_{N-m+1}$\}, where $X_i = \{x_i, x_{i+1},..., x_{i+m-1}\}$. The equation of sample entropy is:

\begin{equation}
    \label{entropy}
    SampEn = - \log\frac{U^m}{U^{m+1}}
\end{equation}
where
\begin{equation}
    U^m = - \frac{1}{N-m}\sum_{i=1}^{N-m}U^m_i
\end{equation}
\begin{equation}
    U^m_i = \frac{[\#\; of\; j\;|d(X_i, X_j) < r]}{N-m-1}
\end{equation}
In our case, the distance $d$ is:
\begin{equation}
    d(X_i, X_j) = \max |X_i - X_j| = \max_{k=1,...,m}|x_{i+k-1} - x_{j+k-1}|
\end{equation}
Generally, $m=2$ and $r$ = [0.2 * (SD of $X$)].

\subsubsection{Sleep} From Fitbit sensors, we obtained sleep duration and sleep efficiency. Then, we calculated the mean and SD values of sleep duration and sleep efficiency across the previous 7, 5, and 3 days. Moreover, using sleep data in one-minute resolution, we calculated sleep regularity with sliding windows across 7 days of participants' data. Sleep regularity is a value of 0 - 1 based on the likelihood of sleep/wake state being the same time-points 24 hours apart, and  is associated with health, wellbeing, and academic performance in college students \cite{sano2015measuring, phillips2017irregular, fischer2020irregular}.
From daily surveys, we obtained a daily feature of the time taken to fall asleep in minutes. Participants also reported how they woke up in the morning: waking up naturally, being awakened by the alarm, or other than alarm.
Naps have been shown a positive impact on shift workers' performance, alertness \cite{purnell2002impact}, and wellbeing \cite{li2019napping}. From participants' questionnaires, we summarized the times and total duration of naps across a day.

\subsubsection{Steps} Total daily number of steps and minute by minute number of steps were recorded in the Fitbit dataset. To measure the variability of participants' physical activities, we computed the mean and SD to indicate step variability across the previous 7, 5, and 3 days. Excluding the sleep time, we counted the minutes of: (i) duration of segments without steps (stationary segments) and (ii) duration of segments with continuous steps (active segments) in 1-min bins. We used the following information entropy equation to calculate the entropy of the two types of physical activity based  stationary and active segments: 
\begin{equation}
    En = - \sum_i p_i \log p_i
\end{equation}
where $p_i$ represents the probability that the $i_{th}$ item was observed.

\subsubsection{Work} Work schedules and work hours are directly related to symptoms such as sleep disorders and chronic fatigue  \cite{bourdouxhe2000interaction}. Also, excessive work hours are harmful to workers' health and wellbeing \cite{golden2008overtime}. We engineered work related features such as daily work shifts, total work duration per day, and overwork duration in minutes according to participants' answers in the questionnaires. There were three different work shifts, and each shift was for eight hours (1: 8:30-16:30, 2: 16:30-0:30, 3:0:30-8:30). Total work duration was actual work time, and the overwork duration was the difference between the actual work hours and the scheduled hours.

\subsubsection{Caffeine, alcohol and drug use} Considering caffeine, alcohol, and drug intake affects workers' alertness \cite{pasman2017effect, roehrs2001sleep}, we computed features related to the intake of caffeinated drinks, drug, and alcohol based on the participants' reports: the number of caffeinated drinks per day, and a binary feature for indicating whether the participant had drug or alcohol each day.

% \textcolor{blue}{From the Fitbit app, we directly get the daily features of sleep duration, sleep efficiency, and steps. Besides, we compute features including daily mean, SD, and entropy of heart rates, because heart rate and heart rate variability are related to work stress  \cite{vrijkotte2000effects} and mood \cite{shapiro2001striking}. \textcolor{red}{explain how you computed the entropy} According to the previous study that sleep irregularity is associated with participant's mood \cite{sano2015measuring}, we extract SD of sleep duration \& sleep efficiency and compute sleep irregularity by checking the differences of sleep status of consecutive two days \textcolor{red}{is this true? this is different from our sleep regularity measure}. We also calculate the SD of steps across the previous 3,5,7 days and the entropy of stepping activities across a day. From the survey of participants, we extract the features of the number and duration of naps they took, as researchers indicate that napping is beneficial to shift working nurses' wellbeing \cite{li2019napping}. Considering the intakes such as caffeine, alcohol, and drug affect the alertness of people \cite{pasman2017effect, roehrs2001sleep}, we make the dosage of coffee and intaking of drugs and alcohol as features. We also use information about participants' sleep conditions, such as sleep onset/offset time, time to fall asleep, and awake type (natural awaking/alarm awaking) as features.}

\subsection{Statistical Analysis Of Physiological And Behavioral Features Between Nurses And Doctors}
%To check the difference of physiological and behavioral patterns
%between two different job roles, nurses and doctors, 
We applied statistical tests to analyze the differences of physiological and behavioral features between two groups, nurses and doctors. Seventy-seven days of data were in the group of doctors, and 164 days of data were in the nurses' group. Between 2 groups, we compared the numeric features such as daily average heart rate, steps, and overwork time using Mann-Whitney U test (non-normally distributed features) \cite{mann1947test} and Welch's t-test (normally distributed features) \cite{welch1947generalization}, whereas the categorical features such as awakening types and working shifts were compared with chi-square test \cite{pearson1900x}. %The results of the statistical analysis shown in section \ref{data_statistics}.

\subsection{Job-role Based Multitask Multilabel Neural Network}
%\textcolor{red}{I still feel this is just multitask leraning but tasks as job-roles and wellbeing labels?}
Neural networks have been widely used in various areas, including face detection \cite{rowley1998neural}, mood, health, and stress prediction \cite{taylor2017personalized}. These previous outstanding works showed that the design of neural network structure needs the consideration of unique characteristics of data sets used in different applications. As discussed briefly in section \ref{intro}, in this work, we considered two important aspects: (1) different distributions in health and wellbeing labels based on our participants' demographic information and (2) correlations among health and wellbeing labels.
We observed differences in the distributions of self-reported health and wellbeing labels from two job roles, nurses and doctors. Also, there are correlations among the five labels. The details of the data statistics will be discussed in section \ref{data_statistics}. 

\begin{figure}
    \centering
    \includegraphics[width=1\linewidth]{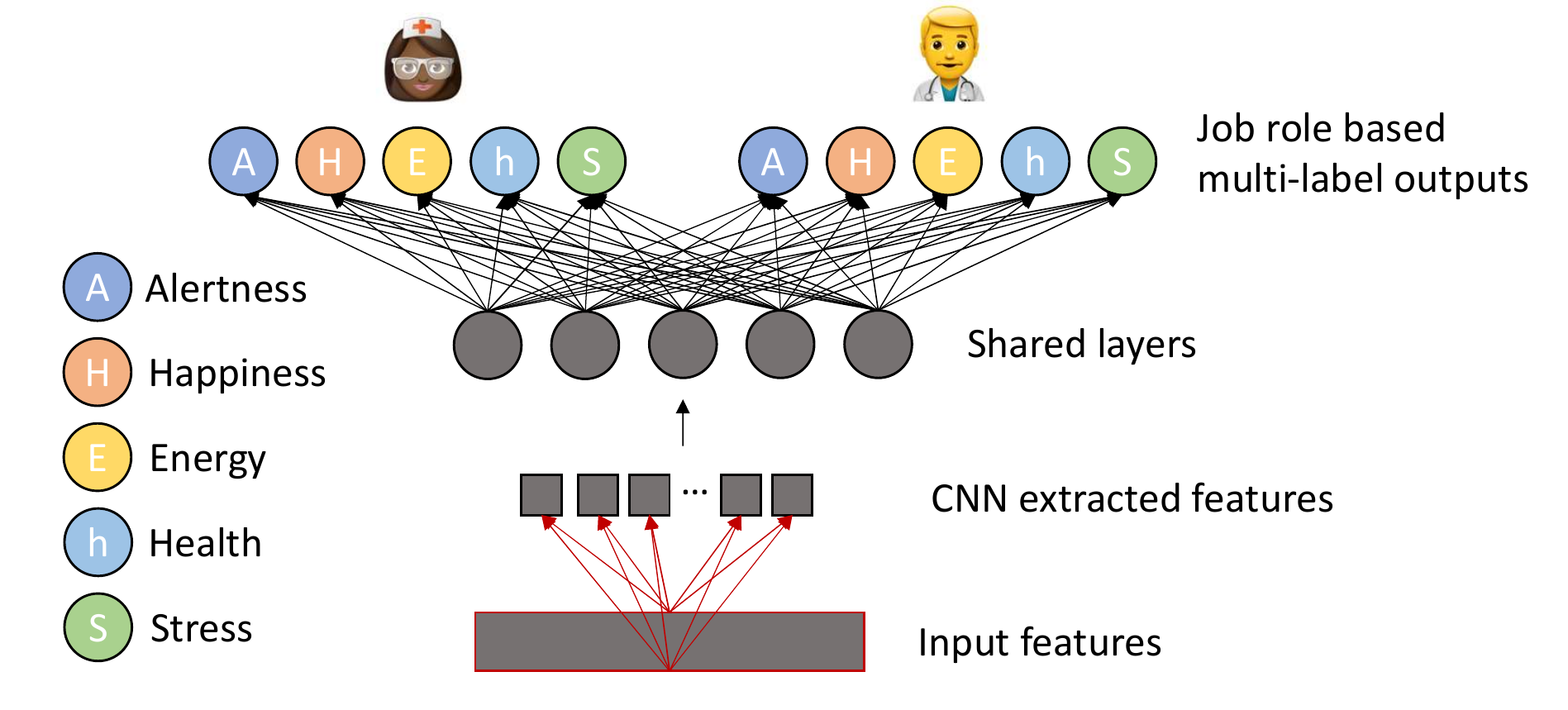}
    \caption{A simplified version of our job-role based multitask multilabel neural network. Convolutional neural network kernels are applied for extracting high-level features. Our health and wellbeing prediction is designed for nurses and doctors using a portion of the network trained only using data from either nurses or doctors. Shared layers learn representation from all participants. The final output layers provide the prediction of all five labels simultaneously. }
    \label{model_structure}
\end{figure}

To learn different representations corresponding to participant job roles, we applied a multitask learning method, which divided tasks according to participants' job roles. Furthermore, as another form of multitask learning, we used multilabel learning for considering different health and wellbeing labels as tasks. In this way, the model would also fit the correlation among labels. In this work, we designed a job-role based multitask and multilabel neural network model that leveraged user demographic information and correlations among labels at the same time. Figure \ref{model_structure} shows a simplified version of our model. When training the model, there might be redundant features in our input data that would not help health and wellbeing prediction. In contrast, some non-linear combinations of features might improve our model performance. Thus, we applied a one-dimension convolutional neural network (CNN) layer to extract auto-features from our inputs. As shown in Figure \ref{model_structure}, we designed convolutional kernels to learn higher-level features across every day feature vectors: 32 row-wise convolutional kernels embedded 32 channels of new features. Then, the CNN extracted features were fed into the multitask neural network. The shared layers in the network learn the representation from all participant data, and the divided branches of the network structure learn the representation independently from participants in different job roles, nurses and doctors. When doctors' data are fed into the model for training, the weights of loss and optimizer of the nurse branch will be set to 0, and vice verse. Furthermore, each branch of the network outputs all five labels (alertness, happiness, energy, health, and stress) from the shared network layers. Therefore, the outputs of our model simultaneously provide the prediction of all five labels for nurses and doctors. The batch loss function of our model can be represented as:

\begin{equation}
    L = \sum_{nurse} L_{ml} + \sum_{doctor} L_{ml}
\end{equation}
\begin{equation}
    L_{ml} = \sum_{l = \{alert, happy, energy, health, stress\}} loss(x, y_l)
\end{equation}
Where $x$ and $y$ represent the input data and the expected output target, respectively. $loss$ is mean squared error loss in regression tasks and cross-entropy loss in the classification tasks.

\section{Experiments}

Our tasks are formulated in two ways for evaluation: regression and classification tasks. The regression task is to predict the next day's health and wellbeing scores, each in the range of 0-100, whereas the classification task is to predict next day's high/low (binary classification, defined as 100-51, 50-0) or high/mid/low health and wellbeing levels, and high/ mid/low (3-class classification, defined as 100-67, 66-34, or 33-0). Our models use the wearable and survey data up to and including the current day for predicting nurses' and doctors' next day health and wellbeing labels. 

%\textcolor{red}{throughout the paper, we need to use consistent terms, sometimes we use multitask or multitask, multilabel, multilabel. Let's decide which terms we use, with hyphen or not?}
We compared our job-role based multitask multilabel model (MTML-NN) with following approaches to evaluate the benefits of using demographic information and the correlation among labels: (1) random forest (RF), (2) RBF kernel based support vector machine (SVM), (3) multitask neural network (MT-NN) that used clusters of participants and achieved state-of-the-art performance in a previous study\cite{taylor2017personalized}, (4) multitask neural network with labels as tasks (ML-NN). In addition to applying ML-NN to all participants (ML-NN (all)), we also calculated the prediction results for nurses (ML-NN(N)) and doctors (ML-NN(D)) separately. 

For training and testing our models, we randomly split the dataset into training and testing data in a ratio of 80\% to 20\%. We applied 10-fold cross-validation and grid search to finalize the hyperparameters for all models mentioned above in the training set. Then,  we tested models in the testing set. To make the evaluation process more robust, we repeated the random data split strategy (training/testing : 80\%/20\%) 10 times to evaluate the model performance. As the evaluating metrics, we use mean absolute errors for the regression models and f1-scores for classification tasks. 
Furthermore, we adopt focal loss\cite{lin2017focal} as the objective function in the classification tasks to mitigate the unbalanced sample size in both binary and 3-class tasks. The Adam optimizer \cite{kingma2014adam} was used in training the neural networks, with a learning rate of 0.005 and 0.9, 0.999 for $\beta_{1}$ and $\beta_{2}$. 

\subsection{Model Weights Analysis}
In addition to the prediction performance, interpretability is also an essential part of machine learning models. Ideally, we would like to provide our prediction results along with reasonable explanations to our participants or health/medical stakeholders. First, from the weights in the RF model, we analyzed the importance of input features.
Then, in our deep learning MTML-NN model, we analyzed the importance of the features by examining the parameters in the first CNN layer before the non-linear activation function. Since the CNN kernel we designed is in one-dimension with a size of the number of features, and parameters in the CNN kernel would correspond to the input features. We calculated the average value of each feature on all channels to check the importance of the features. Also, we computed the correlations between the output of the CNN layer and the input features. Features that have higher correlations with the CNN outputs would also be considered important features.

\section{Results \& Discussion}
\subsection{Data statistics}
\label{data_statistics}
\begin{figure}
    \centering
    \includegraphics[width=.6\linewidth]{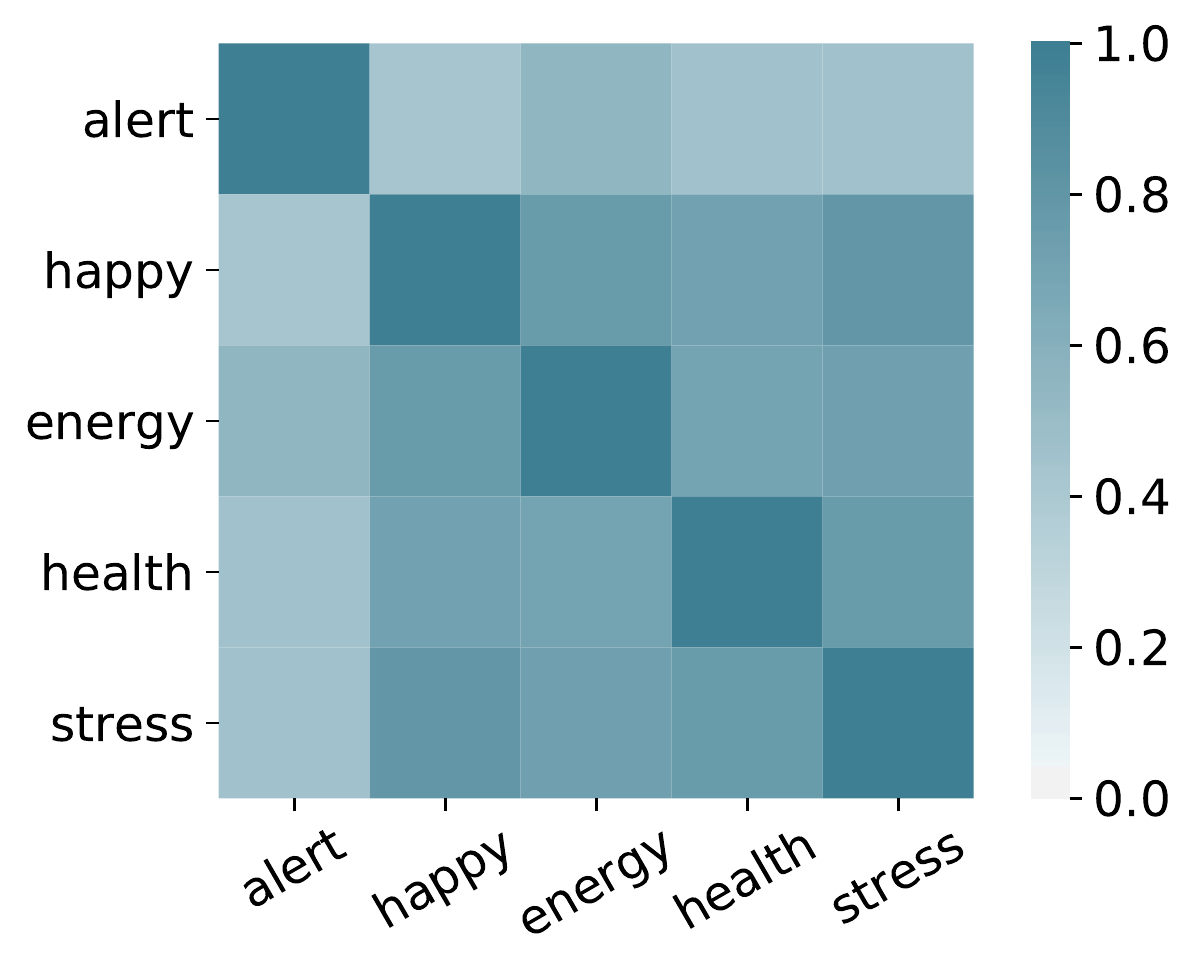}
    \caption{Correlation coefficients matrix of wellbeing labels}
    \label{corr_matrix_labels}
\end{figure}

As shown in Table \ref{label_dist_table}, the average score of alertness label was the lowest among all five labels; while the stress label (0: pressure-1: calm) showed the highest average score. Compared with other labels, the SD of happiness score was lower. Moreover, the distribution of health and wellbeing labels for nurses and doctors were different. For example, doctors generally had higher subjective alertness and energy than nurses in the morning. In addition, we computed correlations among the five health and wellbeing labels. Figure \ref {corr_matrix_labels} shows the correlation coefficients matrix of all labels, and there are different degrees of correlation among the labels. The Pearson test \cite{mukaka2012guide} showed that all five labels were significantly correlated. The linear fitting coefficient of determination ($r^2$) values \cite{nagelkerke1991note} between the alert label and other labels ranged from 0.19 to 0.28, while the $r^2$ values among the happy, energy, health and stress labels were all higher than 0.55, with the highest value being 0.70 (happy and stress).

We also compared feature distributions between nurses and doctors (Table \ref{features_n_d}). We found that the mean heart rate of doctors was significantly higher than that of nurses; whereas the variability of heart rate, defined as SD and sample entropy, was higher in nurses than doctors. In terms of sleep, we found that doctors showed higher sleep efficiency and lower sleep irregularity than nurses. Further, We found statistical differences between nurses and doctors in movement features, including mean and SD of daily steps across the previous 7 days, and the entropy for stationary/active segments. We did not observe any statistical differences between nurses and doctors in working shifts and total work hours among shift work features. However, we found that overwork was more common among doctors.

\begin{table}
    \centering
    \caption{Mean(SD) of daily wellbeing \& P-values from Welch's t-test} 
    \begin{tabular}{c|cc|c}\label{label_dist_table}
                    & Nurse       & Doctor      & p-value \\ \hline
    Alertness & 38.5 (22.9) & 52.8 (23.5) & $<$ 0.05               \\
    Happiness & 57.2 (20.9) & 59.2 (17.1) & 0.38                   \\
    Energy    & 54.0 (22.9) & 60.5 (22.4) & $<$ 0.05               \\
    Health    & 63.3 (22.1) & 63.9 (22.4) & 0.83                   \\
    Stress    & 63.5 (23.4) & 65.6 (17.5) & 0.39                   \\ \hline
    \end{tabular}
    
\end{table}

\begin{table}
\centering
\caption{List of main features and the statistics of nurses and doctors. The statistics of numeric features are shown in mean (SD) values, and the differences are tested with Mann-Whitney U-test (non-normally distributed features) and Welch's t-test (normally distributed features, indicated with $\star$); whereas we use percentages to reveal the statistics of categorical features and apply the chi-square test to check their statistical differences.}
\vspace{0.5cm}
\label{features_n_d}
\begin{tabular}{|c|c|c|c|c|}
\hline
Source                   & Daily Features                                                                                                                 & Nurses (N= 10)                                                                                        & Doctors (N= 4)                                                                                     & P-value  \\ \hline
\multirow{20}{*}{Fitbit} & Heart Rate (HR) - Mean $\star$                                                                                                        & 78.5 (7.1)                                                                                   & 70.6 (6.8)                                                                                 & $<$ 0.05 \\ \cline{2-5} 
                         & Heart Rate(HR) - SD                                                                                                           & 13.5 (3.5)                                                                                   & 12.3 (3.2)                                                                                 & $<$ 0.05 \\ \cline{2-5} 
                         & Heart Rate(HR) - Entropy $\star$                                                                                                      & 0.64 (0.23)                                                                                  & 0.69 (0.20)                                                                                & $<$ 0.05 \\ \cline{2-5} 
                         & Sleep Duration (mins)                                                                                                                & 374.3 (134.0)                                                                                & 363.1 (106.3)                                                                              & 0.36     \\ \cline{2-5} 
                         & Sleep Efficiency (0-100)                                                                                                              & 93.1 (4.9)                                                                                   & 95.5 (2.9)                                                                                 & $<$ 0.05 \\ \cline{2-5} 
                         & Sleep Regularity $\star$                                                                                                               & 0.31 (0.25)                                                                                  & 0.26 (0.19)                                                                                & $<$ 0.05 \\ \cline{2-5} 
                          & \begin{tabular}[c]{@{}c@{}}Sleep Duration \\ - Mean across previous 7 days\end{tabular}                                         & 370.5 (74.4)                                                                                 & 359.0 (46.1)                                                                                & 0.21 \\ \cline{2-5} 
                         & \begin{tabular}[c]{@{}c@{}}Sleep Efficiency \\ - Mean across previous 7 days\end{tabular}                                       & 94.8 (3.3)                                                                                  & 93.2 (3.6)                                                                                & $<$ 0.05     \\ \cline{2-5} 
                         & \begin{tabular}[c]{@{}c@{}}Sleep Duration \\ - SD across previous 7 days\end{tabular}                                         & 106.8 (39.1)                                                                                 & 88.4 (37.8)                                                                                & $<$ 0.05 \\ \cline{2-5} 
                         & \begin{tabular}[c]{@{}c@{}}Sleep Efficiency \\ - SD across previous 7 days\end{tabular}                                       & 2.19 (1.31)                                                                                  & 2.20 (1.02)                                                                                & 0.21     \\ \cline{2-5} 
                         & Steps $\star$                                                                                                                          & 8931.2 (4030.3)                                                                              & 8139.9 (3350.8)                                                                            & 0.21     \\ \cline{2-5} 
                         & \begin{tabular}[c]{@{}c@{}}Steps \\ - Mean across previous 7 days\end{tabular} $\star$                                                  & 8684.8 (2372.8)                                                                              & 9063.4 (2236.9)                                                                             & $<$ 0.05 \\ \cline{2-5} 
                         & \begin{tabular}[c]{@{}c@{}}Steps \\ - SD across previous 7 days\end{tabular} $\star$                                                  & 3099.5 (1017.8)                                                                              & 2582.5 (849.7)                                                                             & $<$ 0.05 \\ \cline{2-5} 
                         & Entropy (stationary segments)                                                                                                     & 2.17 (0.53)                                                                                  & 2.61 (0.24)                                                                                & $<$ 0.05 \\ \cline{2-5} 
                         & Entropy (active segments)                                                                                                   & 1.67 (0.45)                                                                                  & 1.65 (0.19)                                                                                & $<$ 0.05     \\ \hline
\multirow{20}{*}{Survey}  & Number of Naps                                                                                                                 & 0.55 (0.68)                                                                                  & 0.37 (0.66)                                                                                & $<$ 0.05 \\ \cline{2-5} 
                         & Duration of Naps (mins)                                                                                                        & 31.1 (68.4)                                                                                  & 19.1 (49.4)                                                                                & 0.12     \\ \cline{2-5} 
                         & \# of Cups of Caffeinated Drinks                                                                                                  & 0.47 (0.73)                                                                                  & 0.45 (0.62)                                                                                & 0.45     \\ \cline{2-5} 
                         & \begin{tabular}[c]{@{}c@{}}Wake-up Type \\ - Natural \\ - Alarm \\ - Other than alarm \end{tabular}                              & \begin{tabular}[c]{@{}c@{}}-\\ 35.5\%\\ 60.9\%\\ 3.6\%\end{tabular}                          & \begin{tabular}[c]{@{}c@{}}-\\ 35.6\%\\ 46.5\%\\ 17.8\%\end{tabular}                       & $<$ 0.05 \\ \cline{2-5} 
                         & \begin{tabular}[c]{@{}c@{}}Time to Fall Asleep (mins)\\ - 0-5 \\ - 6-15 \\ - 16-30 \\ - 31-45 \\ - 45-60\\ - 60+ \end{tabular} & \begin{tabular}[c]{@{}c@{}}-\\ 34.0\%\\ 31.4\%\\ 17.3\%\\ 6.6\%\\ 3.6\%\\ 7.1\%\end{tabular} & \begin{tabular}[c]{@{}c@{}}-\\ 40.6\%\\ 41.6\%\\ 11.9\%\\ 3.0\%\\ 3.0\%\\ 0\%\end{tabular} & $<$ 0.05 \\ \cline{2-5} 
                         & \begin{tabular}[c]{@{}c@{}}Work Shifts \\ - Shift 1 \\ - Shift 2 \\ - Shift 3\end{tabular}                                   & \begin{tabular}[c]{@{}c@{}}-\\ 53.8\%\\ 30.4\%\\ 15.7\%\end{tabular}                         & \begin{tabular}[c]{@{}c@{}}-\\ 64.3\%\\ 19.8\%\\ 15.8\%\end{tabular}                       & 0.12     \\ \cline{2-5} 
                         & Work Time (hours)                                                                                                                    & 8.0 (0.0)                                                                                    & 8.4 (1.7)                                                                                  & 0.08     \\ \cline{2-5} 
                         & Overwork Time (mins)                                                                                                        & 11.0 (41.8)                                                                                  & 202.6 (320.1)                                                                              & $<$ 0.05 \\ \hline
\end{tabular}
\end{table}

\subsection{Wellbeing prediction}

The classification and regression performance using different models is shown in Table \ref{perforamnce_table}. Our proposed job-role based MTML-NN performed the best for four labels in binary classification and all wellbeing labels in 3-class classification and regression (ANOVA, Tukey, p $<$ 0.05). Our results showed the benefits of our proposed simultaneous job role and correlated label modeling, especially in 3-class classification and regression. However, according to the performance in 3-class classification, we found poor classification performance for some classes. For example, in the 3-class alertness classification, the high-alertness class precision and recall values were only 0.16 and 0.27 in respectively; and our low-energy class prediction  was also relatively low with a precision of 0.33 and a recall of 0.24. These errors might come from the data imbalance problem. In the 3-class classification tasks, the high alertness labels accounted for only 20\% of all labels, and the low energy labels accounted for 15\% of all labels.

\begin{table}
  \centering
  \caption{Prediction performance (f1-score for classification; mean absolute error (MAE) for regression) of different algorithms. Bold entries represent statistically significantly better results over the other models.}
  \begin{tabular}{c|c|ccccc} \label{perforamnce_table}
Tasks                       & Algorithms  & Alertness              & Happiness             & Energy                & Health                & Stress                \\ \hline
\multirow{8}{*}{Binary}     & RF          & 50\%$\pm$7\%           & 78\%$\pm$4\%          & 65\%$\pm$4\%          & \textbf{84\%$\pm$3\%}          & 82\%$\pm$3\%          \\
                            & SVM         & 52\%$\pm$4\%           & 69\%$\pm$4\%          & 62\%$\pm$6\%          & 80\%$\pm$4\%          & 77\%$\pm$5\%          \\
                            & NN          & 55\%$\pm$4\%           & 71\%$\pm$5\%          & 65\%$\pm$3\%          & 82\%$\pm$3\%          & 80\%$\pm$3\%          \\
                            & MT-NN       & 60\%$\pm$4\%           & 76\%$\pm$3\%          & 69\%$\pm$5\%          & \textbf{83\%$\pm$4\%}          & \textbf{83\%$\pm$4\%}          \\
                            & ML-NN (all) & 55\%$\pm$7\%           & 74\%$\pm$7\%          & 68\%$\pm$4\%          & 80\%$\pm$4\%          & \textbf{83\%$\pm$3\%}          \\
                            & ML-NN (N)   & 55\%$\pm$9\%           & 69\%$\pm$5\%          & 64\%$\pm$6\%          & 79\%$\pm$7\%          & 79\%$\pm$7\%          \\
                            & ML-NN (D)   & 59\%$\pm$8\%           & 75\%$\pm$5\%          & 67\%$\pm$7\%          & \textbf{85\%$\pm$5\%}          & \textbf{85\%$\pm$7\%}          \\
                            & MTML-NN     & \textbf{64\%$\pm$7\%}  & \textbf{79\%$\pm$3\%} & \textbf{71\%$\pm$4\%} & 81\%$\pm$3\%          & \textbf{84\%$\pm$3\%}          \\ \hline
\multirow{8}{*}{3-class}    & RF          & 53\%$\pm$5\%           & 39\%$\pm$5\%          & 46\%$\pm$6\%          & 49\%$\pm$7\%          & 44\%$\pm$5\%          \\
                            & SVM         & 47\%$\pm$7\%           & 40\%$\pm$5\%          & 43\%$\pm$5\%          & 49\%$\pm$7\%          & 46\%$\pm$4\%          \\
                            & NN          & 51\%$\pm$5\%           & 45\%$\pm$6\%          & 45\%$\pm$5\%          & 53\%$\pm$6\%          & 51\%$\pm$4\%          \\
                            & MT-NN       & 57\%$\pm$6\%           & 46\%$\pm$5\%          & 48\%$\pm$6\%          & 53\%$\pm$5\%          & 50\%$\pm$5\%          \\
                            & ML-NN (all) & 52\%$\pm$8\%           & \textbf{53\%$\pm$7\%}          & 48\%$\pm$7\%          & 55\%$\pm$3\%          & 54\%$\pm$4\%          \\
                            & ML-NN (N)   & 45\%$\pm$7\%           & \textbf{54\%$\pm$5\%}          & 49\%$\pm$5\%          & 56\%$\pm$2\%          & 53\%$\pm$5\%          \\
                            & ML-NN (D)   & 54\%$\pm$5\%           & \textbf{51\%$\pm$7\%}          & 45\%$\pm$6\%          & 54\%$\pm$3\%          & 52\%$\pm$7\%          \\
                            & MTML-NN     & \textbf{59\%$\pm$5\%}  & \textbf{52\%$\pm$4\%}          & \textbf{51\%$\pm$4\%} & \textbf{58\%$\pm$7\%} & \textbf{57\%$\pm$5\%} \\ \hline
\multirow{7}{*}{Regression} & SVR         & 20.6$\pm$2.9           & 19.7$\pm$2.5          & 21.3$\pm$2.4          & 18.9$\pm$1.8          & 21.7$\pm$2.1          \\
                            & NN          & 19.9$\pm$1.8           & 19.0$\pm$1.9          & 20.3$\pm$2.2          & 19.5$\pm$1.9          & 20.4$\pm$1.9          \\
                            & MT-NN       & 19.4$\pm$2.1           & 18.8$\pm$2.3          & 20.7$\pm$1.6          & 19.3$\pm$2.0          & 20.3$\pm$1.7          \\
                            & ML-NN (all) & 18.6$\pm$1.3           & 16.9$\pm$3.1          & 18.6$\pm$2.0          & 18.7$\pm$2.6          & 19.4$\pm$2.6          \\
                            & ML-NN (N)   & 18.0$\pm$1.1           & 15.9$\pm$1.9          & 17.3$\pm$1.7          & 15.6$\pm$1.7          & 17.7$\pm$1.3          \\
                            & ML-NN (D)   & 20.4$\pm$2.1           & 16.0$\pm$2.0          & 19.3$\pm$2.0          & 17.4$\pm$2.0          & 19.4$\pm$3.1          \\
                            & MTML-NN     & \textbf{17.4 $\pm$1.4} & \textbf{15.1$\pm$1.6} & \textbf{17.7$\pm$1.2} & \textbf{15.4$\pm$1.5} & \textbf{15.6$\pm$1.9} \\ \hline
\end{tabular}
\end{table}

Furthermore, we also found the benefits of using job role information or multiple labels separately. For example, in the alertness prediction, job-role based MT-NN showed significant improvement from NN for both binary and 3-class classification. Besides the overall f1-score, we observed some improvements revealed in each class. For example, in the 3-class alertness classification tasks, MT-NN model provided significantly higher recall and precision scores in low and middle alertness classification compared to NN model (Welch's t-test, p $<$ 0.05). We did not observe any significant improvement in the regression tasks. However, the average prediction MAE of MT-NN was lower than that of NN. Significant improvements were observed in ML-NN compared to NN in almost all tasks. For example, in the regression tasks, the ML-NN (all) performed statistically significantly better than NN in predicting alertness, happiness, energy, and stress labels.

\subsection{Weight Analysis}
From the RF model, for both the binary and 3-class classification happiness prediction tasks, we found features including mean heart rate and heart rate sample entropy across the day, sleep duration, sleep regularity, and the SD of sleep efficiency across the previous seven days, were the most important. In the alertness prediction tasks, work shifts, stationary segment entropy, mean step, and mean sleep duration across the previous 7, 5 days played important roles. 
The analysis of the parameters in the CNN layer in the MTML-NN model and the correlations between the CNN output and input features indicated that features including heart rate sample entropy, sleep regularity, sleep efficiency, work shifts, steps, and active segment entropy - contributed to health and wellbeing prediction. For example, from the correlation analysis, we found that the sleep efficiency, sleep regularity, and daytime work shift were positively related to the wellbeing (Pearson test, p-value $<$ 0.05 / (\# of features)); whereas the step and the entropy of active segments were negatively related to the wellbeing (Pearson test, p-value $<$ 0.05 / (\# of features)).
Our findings were consistent with some prior results. For example, according to the previous works, sleep influences physical and psychological health \cite{wong2013interplay}, and stress \cite{mezick2009intra}; sleep regularity is associated with mood \cite{sano2015measuring}. Previous studies also indicated the association between work shifts and stress levels \cite{wisetborisut2014shift}.

\section{Conclusion}
In this work, we collected physiological and behavioral wearable sensor data as well as survey data from shift-work nurses and doctors, and compared their physiology and behaviors between two job roles. Then, we proposed a job-role based multitask and multilabel learning model structure to predict shift workers' health and wellbeing for next day using sensor and questionnaire data. The proposed model outperformed the benchmark models, including RF and SVM as well as the previous state-of-the-art models. The analysis of model weights showed that health rate, work shifts, sleep parameters such as sleep regularity and sleep efficiency contributed to shift workers' health and wellbeing labels. As future work, we will collect more data from shift workers and design a system to improve shift workers' health and wellbeing. 

\bibliographystyle{splncs04}
\bibliography{references}
\end{document}